\documentclass[10pt,twocolumn,letterpaper]{article}

\usepackage[numbers]{natbib}
\usepackage{wacv}
\usepackage{times}
\usepackage{epsfig}
\usepackage{graphicx}
\usepackage{amsmath}
\usepackage{amssymb}

\usepackage{xcolor}
\usepackage{amsthm}
\usepackage{subcaption}
\usepackage{graphicx}

\def\wacvPaperID{735} 

\wacvfinalcopy 

\ifwacvfinal
\def\assignedStartPage{1} 
\fi

\ifwacvfinal
\usepackage[breaklinks=true,bookmarks=false]{hyperref}
\else
\usepackage[pagebackref=true,breaklinks=true,colorlinks,bookmarks=false]{hyperref}
\fi

\usepackage{cleveref}

\ifwacvfinal
\else
\fi

\theoremstyle{definition}
\newtheorem{definition}{Definition}

\theoremstyle{definition}
\newtheorem{hypothesis}{Hypothesis}

\newcommand{\softmax}{softmax}
\newcommand{\ce}{CrossEntropy}

\newcommand{\david}[1]{\emph{\textcolor{blue}{(David: #1)}}}

\renewcommand{\david}[1]{}

\begin{document}

\title{Conflicting Bundles: Adapting Architectures Towards the Improved  \\ Training of Deep Neural Networks}

\author{David Peer \qquad
Sebastian Stabinger \qquad
Antonio Rodr\'{i}guez-S\'{a}nchez\\
University of Innsbruck \\
Austria\\
{\tt\small https://iis.uibk.ac.at/}
}

\maketitle

\begin{abstract}
Designing neural network architectures is a challenging task and knowing which specific layers of a model must be adapted to improve the performance is almost a mystery. In this paper, we introduce a novel theory and metric to identify layers that decrease the test accuracy of the trained models, this identification is done as early as at the beginning of training. In the worst-case, such a layer could lead to a network that can not be trained at all. More precisely, we identified those layers that worsen the performance because they produce conflicting training bundles as we show in our novel theoretical analysis, complemented by our extensive empirical studies. Based on these findings, a novel algorithm is introduced to remove performance decreasing layers automatically. Architectures found by this algorithm achieve a competitive accuracy when compared against the state-of-the-art architectures. While keeping such high accuracy, our approach drastically reduces memory consumption and inference time for different computer vision tasks.
\end{abstract}

\section{Introduction}\label{sec:introduction}

\begin{figure}[t]
  \centering
  \captionsetup{justification=centering}
  \includegraphics[width=.71\linewidth]{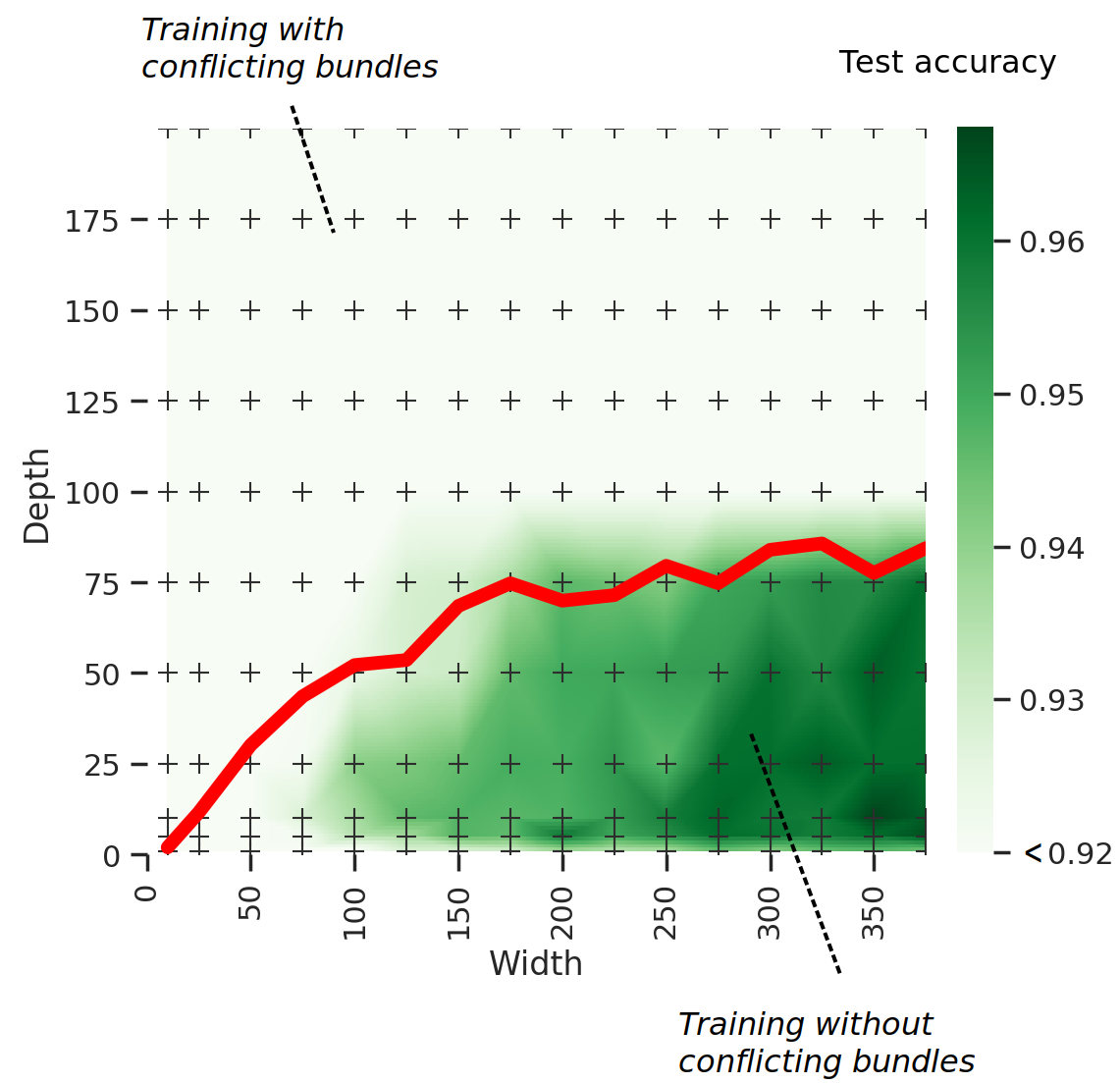}
  \caption{Test accuracy of different neural networks trained on Mnist together with the boundary that separates networks that induce conflicting training bundles. }\label{fig:experiment_fnn_first_conflicting_layer}
\end{figure}

The training of deep neural networks is a complex and challenging task \cite{unsupervised_pretraining}, one that can be achieved by better initialization strategies \cite{glorot_initializer, he_initializer}, activation functions \cite{relu, mish}, regularization methods \cite{batch_normalization, dropout} and network architectures \cite{residual_neural_networks, gammacapsules, highway_networks, efficientnet}. Designing good architectures is still a mystery and often done through trial and error approaches or exhaustive grid searches. A test accuracy map is shown in \cref{fig:experiment_fnn_first_conflicting_layer} for $176$ different fully connected neural networks trained on Mnist. The expressivity of networks grows exponentially with depth \cite{capsnet_limitations, expressive_power}, but it can be seen that the optimal architecture w.r.t test accuracy that is also trainable depends on a good balance between width and depth \cite{efficientnet}. Searching through many different architectures this way to find a good configuration is not an option for real-world applications as too many possible configurations exist and the time required to train and evaluate each alternative is usually too long \cite{how_many_layers_and_nodes}. We introduce the \emph{conflicting training bundle problem} in this paper and we can prove theoretically and experimentally that conflicting bundles worsen the performance of neural network models. The red line in  \cref{fig:experiment_fnn_first_conflicting_layer} is a boundary that separates all models that are trained without conflicting bundles (below the line) and all models that induce conflicting bundles during training (above the line). We are able to locate layers that induce conflicting bundles already at the beginning of the training by evaluating only the deepest network for each width as we motivate later in the paper. 
Therefore, our method to find this boundary is computationally very cheap and the analysis of conflicting bundles will (1) help researchers to create new or improve state-of-the-art network architectures (2) help to create novel neural architecture search (NAS) algorithms and (3) explain effects from a new theoretical perspective and hopefully inspire new research. The problem that we discovered is described next.

\subsection{The conflicting training bundle problem}
\david{[Done] Motivate why this could occur w.r.t resolution of float}
The output of a neural network is calculated by successively propagating inputs forward through all hidden layers. Those calculations are executed on a CPU or GPU and therefore, all output values of a layer are represented with finite resolution such as \texttt{float32}. Consequently, two outputs that are only slightly different (less than the minimum resolution of float), are equal on a CPU or GPU. We found that weights are adjusted in wrong directions during training, leading to a worsened overall performance of the model, if any hidden layer produces the same output vector for two input examples with different labels. Therefore, we call two samples \emph{bundled} if the same output w.r.t. the floating point resolution is produced for both inputs after passing through some hidden layer of the network and \emph{conflicting} if both samples are labeled differently. The layer that bundles the samples, called the \emph{conflicting layer}, can be precisely determined and therefore architectures can either manually or automatically be adapted as we demonstrate in this paper.

\subsection{Outline}
In section \ref{sec:theory}, the theory of conflicting training bundles is introduced. We show theoretically that the accuracy decreases if conflicting bundles appear during training and that in the worst-case scenario the network cannot learn from data. In that same section, we introduce a novel metric to quantify and to detect the precise layer that produces conflicting training bundles. In the experimental \cref{sec:experimental} we evaluate many different types of networks: Under tightly controlled settings conflicting training bundles are produced, and the effects are studied. Then, the evolution of conflicting bundles is studied for each training epoch and layer. Fully connected networks, VGG nets and ResNets are tested on different datasets. Finally, a novel NAS algorithm to tackle conflicting training bundles is introduced and compared against the best ResNet found through a grid search. A discussion and inspiration for future research are given in \cref{sec:discussion}.

\subsection{Related work}\label{sec:related_work}
Different optimization problems have already been the subject of study. \citet{efficient_backprop} has shown that careful initialization of the weights of neural networks has a significant effect on the training process. Methods to initialize weights avoid vanishing information during forward-propagation and also avoid vanishing- or exploding gradients during backward propagation \cite{glorot_initializer, he_initializer, unsupervised_pretraining, data_dependent_init, need_good_init}. Very deep networks are difficult to optimize, even when variance-preserving initialization methods are used \cite{highway_networks}. Solutions to overcome this problem include highway networks \citep{highway_networks} that allow for the unimpeded information flow across several layers, or residual learning  \citep{residual_neural_networks} among others. Historical developments and optimization problems that can occur during the training of neural networks are extensively described and summarized by the seminal survey of \citet{deep_learning_in_nn_overview} and the book of \citet{deep_learning_book}.

\david{[Done] Clearify this} To the best of our knowledge, we are the first that precisely locate and quantify the conflicting training bundles problem. Nevertheless, a closely related work published by \citet{vanishing_information} describes that information of original input patterns is lost in higher layers by going through multiple layer transformations and compressions. They also claimed that convolutional neural networks do not suffer from this problem probably due to the high dimensionality of hidden features. In this paper, we study layers in the network hierarchy where as much information vanishes such that two different inputs are represented equally when passed through this layer. We also show that this problem indeed arises for convolutional neural networks that distinguish our paper clearly from the work of \citet{vanishing_information}.

Another closely related work is from \citet{shattered_gradients} which proved that the correlation between gradients in fully connected networks with $L$ layers decay exponentially with $1/2^L$. The correlation decreases only with $1 / \sqrt{L}$ if residual connections are used. Therefore, they concluded that residual networks are easier to train than fully connected networks. Shattered gradients are independent of the width (theorem 1 of \cite{shattered_gradients}) and therefore, the pattern shown in \cref{fig:experiment_fnn_first_conflicting_layer} can not be fully described. On the other hand, we show in this paper that conflicting training bundles depend on the depth and the width. To ensure that conflicting bundles do not shutter gradients, we directly evaluate gradients in \cref{sec:experimental_fully}. The fact that conflicting bundles occur for very shallow networks rules out the vanishing- or exploding gradients \cite{lstm} and we conclude that this is a new type of problem. We also want to mention that related work mainly analyses gradients. Conflicting bundles are analyzed after each layer during forward-propagation which allows to precisely detect the layer that induces the problem.

\section{A theory of conflicting training bundles}\label{sec:theory}

In this section, our theory of conflicting training bundles is introduced. The focus is on classification problems for simplicity, although the idea of conflicting training bundles can also be applied to e.g. regression problems. Let's consider a training set $\mathcal{S} \in \mathcal{X} \times \mathcal{Y}$ that contains objects from a specific domain $\mathcal{X}$ along with its labels $\mathcal{Y}$. Labels $y \in \mathcal{Y}$ are one-hot encoded and the dimensionality of labels is $N_c$. If the input to a layer of a neural network is of dimension $m$ and output of dimension $n$, we use weight matrices $W \in \mathbb{R}^{n \times m}$ and bias terms $b \in \mathbb{R}^{n \times 1}$ to calculate the output. The output of a specific layer $l \in \{1, ..., L\}$ for a network with $L$ layers is a vector $a^{(l+1)}(x_i) = f \left({W}^{(l)} {a}_i^{(l)} + {b}^{(l)} \right)$ for some input $x_i \in \mathcal{X}$ and nonlinearity $f$. We write $a^{(l+1)}_i$ instead of $a^{(l+1)}(x_i)$ and $a^1_i = x_i$. We train neural networks using gradient descent with batches $\mathcal{B} \subseteq \mathcal{X}$ and assume without loss of generality that batches are uniformly distributed w.r.t. the class labels \footnote{This assumption can be easily removed by shifting derivations.}. The loss of the neural network for input $x_i$ is calculated with $J_i = \ce(h_i, y_i)$ and ${h_i} = \softmax({W}^{(L)} {a}_i^{(L)} + {b}^{(L)})$.
The gradient is calculated with $\frac{\partial J_i}{\partial {W}^{(L)}} = ({h_i} - {y_i}) {{a}^{(L)}}^T$ and for a mini-batch with $\frac{\partial J}{\partial {W}^{(L)}} = \frac{1}{|\mathcal{B}|} \sum^{|\mathcal{B}|}_{i=1} \frac{\partial J_i}{\partial {W}^{(L)}}$. We now define for which cases two samples are considered to be in the same bundle and under which circumstances we call a bundle or a layer a \textit{conflicting} one:

\begin{definition} \label{def:bundling}
    Two samples $x_i, x_j \in \mathcal{X}$
    are \emph{bundled} by layer $l$, if $a^{(l+1)}_i = a^{(l+1)}_j$ for the current configuration of learnable parameters.
\end{definition}
\begin{definition} \label{def:conflicts}
    Two samples $x_i, x_j \in \mathcal{X}$
    are \emph{conflicting}, 
    if $y_i \neq y_j$ and there exists some layer $l \in \{1, ..., L\}$ that bundles $x_i$ and $x_j$. We call layer $l$ the \emph{conflicting layer} for $x_i$ and $x_j$.
\end{definition}
Two different input examples $x_i$ and $x_j$ are therefore bundled, if the same output is produced for both inputs after the non-linearity of some hidden layer $l$ is applied. The output layer is per definition not included.
\begin{definition} \label{def:bundle}
    A \emph{bundle} $D_i^l$ contains all samples $x \in \mathcal{B}$ that are bundled after layer $l$ and the set of all bundles $\mathcal{D}^l = \{D^l_1, D^l_2, ... D^l_k\}$ with $1 \leq k \leq |\mathcal{B}|$ ensure that $\forall i \neq j: D^l_i \cap D^l_j = \emptyset$ and $D^l_1 \cup D^l_2, ... \cup D^l_k = \mathcal{B}$.
    A bundle that contains conflicting samples  is called a \emph{conflicting bundle}. We call the bundle $D^l_1$ of the special case $\mathcal{D}^l = \{D^l_1\}$ a \emph{fully conflicting bundle}.
\end{definition}
Per definition, a \emph{conflicting bundle} contains at least two samples with different labels. A \emph{fully conflicting bundle} contains all samples of a mini-batch and therefore all samples are conflicting. We will now make use of the definitions to derive the negative effects that arise when neural networks are trained with fully conflicting bundles or conflicting bundles in general.

\subsection{Fully conflicting bundle}\label{sec:theory_fully}
At first we evaluate the gradient w.r.t. $W^{(L)}$ if a fully conflicting bundle (\cref{def:bundle}) occurs during training:
\begin{align*}
    \frac{\partial J}{\partial {W}^{(L)}} &= \frac{1}{|\mathcal{B}|} \sum^{|\mathcal{B}|}_{i=1} \frac{\partial J_i}{\partial {W}^{(L)}}
    = \frac{1}{|\mathcal{B}|} \sum^{|\mathcal{B}|}_{i=1} \left({h_i} - {y_i} \right) {{a_i}^{(L)}}^T  \\
    &= \left(h -  \frac{1}{N_c} \pmb{1} \right) {{a}^{(L)}}^T
\end{align*}
where $\pmb{1}$ is a vector of dimension $N_c \times 1$ with components $1$. First of all, we can see that all $h_i$ are collapsed into a single $h$, which is the fully conflicting bundle assumption. All labels can also be collapsed into a single $\pmb{1}$ vector scaled by $|B| / N_c$ as we assume uniformly distributed batches. The first observation is that the labels $y_i$ disappeared from the gradient, i.e. the gradient is uncorrelated of $y_i$ and therefore we conclude that the network cannot learn from data. This is also a difference to shattered gradients \cite{shattered_gradients} where gradients become uncorrelated from its input $x_i$ (and not $y_i$). Another observation is that the gradient becomes zero whenever all components of $h$ equal $1 / N_c$ and therefore we hypothesize:

\begin{hypothesis}\label{hyp:fully}
If a fully conflicting bundle occurs during training, all labels are ignored and weights are adjusted until each output neuron fires with constant value $\frac{1}{N_c}$.
\end{hypothesis}

Note that the value of each neuron is $\frac{1}{N_c}$ because we assumed equally distributed labels in the batches. If we relax this assumption, each neuron will fire with a constant value that represents the imbalance of the dataset. For example, if $75\%$ of the examples are of class one, the corresponding neuron will fire with a constant value of $0.75$.

\subsection{Conflicting bundles} \label{sec:theory_partially}
We now relax the assumption that a \emph{fully} conflicting bundle occurs and study the gradient if some conflicting bundles occur (see \cref{def:bundle}):
\begin{align*}
    \frac{\partial J}{\partial {W}^{(L)}} &= 
    \frac{1}{|\mathcal{B}|} \sum_{D \in \mathcal{D}^L} \left( |D| \ h_D -  \sum_{i=1}^{|D|} y_i \right) {{a_D}^{(L)}}^T \\
    &= \frac{1}{|\mathcal{B}|} \sum_{D \in \mathcal{D}^L} \left( |D| \ h_D -  \hat{y}_D \right) {{a_D}^{(L)}}^T
\end{align*}
Compared to the fully conflicting bundle case, where labels disappeared, labels are still included in the calculation of the gradient. But it can also be seen that for the same output $h_D$ different labels are grouped into a single $\hat{y}_D$. Therefore, we assume that the effect of conflicting bundles is similar to the effect of noisy labels for which it is well known that the model performance is worsened \cite{label_noise}. This leads us to the following hypothesis that will be extensively evaluated in the experimental \cref{sec:experimental}:

\begin{hypothesis}\label{hyp:partially}
If conflicting bundles occur during training the performance of the trained model is worsened.
\end{hypothesis}

To check the correctness of \cref{hyp:fully} and \cref{hyp:partially}, we have to be able to detect situations where conflicting bundles occur.
Therefore, a metric to quantify conflicting bundles is introduced next.


\subsection{Conflicting bundle metric}
To measure whether two samples $x_i$ and $x_j$ are bundled by layer $l$ we must check if $a_i^{(l+1)} = a_j^{(l+1)}$ (\cref{def:bundling}). As already mentioned the finite resolution of floating point values must be considered when calculated on a GPU or CPU. Therefore, it makes a difference whether the vectors are used during forward- or backpropagation: During backpropagation, $a_i$ vectors are scaled by the learning rate $\alpha$ and $1 / |\mathcal{B}|$ before they are subtracted from the weights. Therefore, it is possible that very small values that are different during forward propagation are bundled during backpropagation due to the finite resolution of floating-point values. To consider this, we approximate \cref{def:bundling} with
\begin{align} \label{eq:bundled}
    \frac{\alpha}{|\mathcal{B}|} \ ||a_i^{(l+1)} - a_j^{(l+1)}||_\infty \leq \gamma
\end{align}
where $\gamma$ is the currently smallest possible resolution that is supported by the floating-point representation of the GPU or CPU ($\gamma \approx 10^{-8}$). Note that \cref{eq:bundled} depends only on the output of the layer such that this metric can be used for any type of layers such as fully connected, convolutional, pooling or others.

A bundle $D_i^l$ (\cref{def:bundle}) is then the set of all vectors that are equal accordingly to \cref{eq:bundled}. We are mainly interested in \emph{conflicting} bundles and to quantify how conflicting a single bundle ${D_i} \in \mathcal{D}^l$ at training step $t$ and layer $l$ is, we make use of the entropy $H^l(t, D_i)$ with:
\begin{align}
    H^l(t, D_i) = -\sum_{n=1}^{N_c} p(n) \ln \left( p(n) + \epsilon \right)
\end{align}
where $p(n)$ represents the probability that samples of class $n$ occur in bundle $D_i$ and $\epsilon$ is an arbitrarily small value ensuring numerical stability. The entropy $H^l(t, D_i)$ is large if bundle $D_i$ created by layer $l$ contains many examples with different labels and $H^l(t, D_i)$ is small if examples have the same label.

To measure the entropy of all bundles, the size of each bundle must be also considered, because a large conflicting bundle $D_1$ affects the training more than a small conflicting bundle $D_2$. Therefore, we consider the bundle size of each bundle to report the \emph{bundle entropy at training step $t$ for layer $l$} with:
\begin{align}
    H^l(t) = \frac{1}{|\mathcal{B}|} \sum_{D_i\in\mathcal{D}^l} |D_i| \cdot H(t, D_i)
\end{align}
If only one sample or samples of the same class are included in a bundle, $H^l(t)$ is zero. If a fully conflicting bundle occurs then $H^L = \ln(N_c)$.

To be able to check the correctness of \cref{hyp:fully} and \cref{hyp:partially} we must detect whether conflicts occur during training. Therefore, we evaluate $H^l(t)$ after multiple time steps and report an average value of the bundle entropy that occurred during training. We call this metric the \emph{bundle entropy} $H^l$. If not explicitly mentioned we call $H^L$ of the last hidden layer $L$ the bundle entropy. It is important to mention that the bundle entropy $H^L > 0$ iff at least two samples with different labels are bundled during training.

\section{Experimental evaluation}\label{sec:experimental}

In \cref{sec:theory} we hypothesized that if conflicts occur, the test loss increases (\cref{hyp:partially}) and in the extreme case all labels are ignored (\cref{hyp:fully}). In this section, we evaluate whether conflicting bundles occur during the training of SOTA methods and if the effects are as hypothesized.

\subsection{Setup} \label{sec:experimental_setup}

The theory of conflicting bundles should be a new starting point for research into several directions to improve the state-of-the-art, but the focus of this work is to introduce and shed light on the problem itself and not to fine-tune state-of-the-art methods or architectures. Therefore, we suggest the following setup as a starting-point for discussions:

\begin{figure*}[t]
\begin{subfigure}{.33\textwidth}
  \centering
  \captionsetup{justification=centering}
  \includegraphics[width=.99\linewidth]{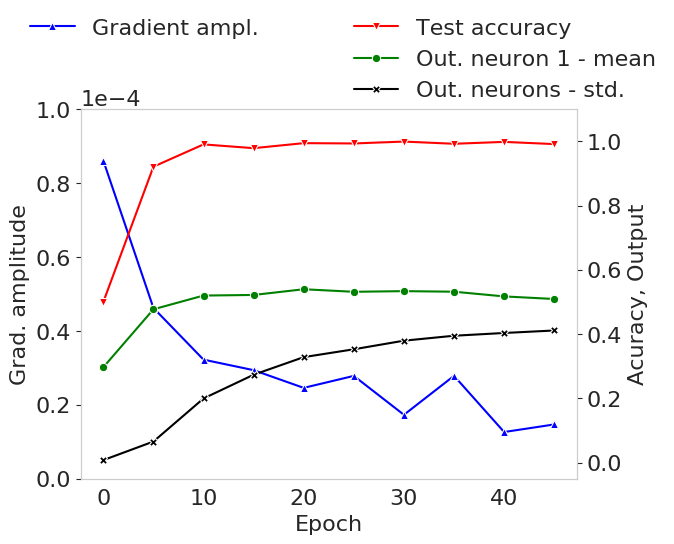}  
  \caption{
  No conflicting bundle \\ balanced dataset.}\label{fig:experiment_threshold_a}
\end{subfigure}
\begin{subfigure}{.33\textwidth}
  \centering
  \captionsetup{justification=centering}
  \includegraphics[width=.99\linewidth]{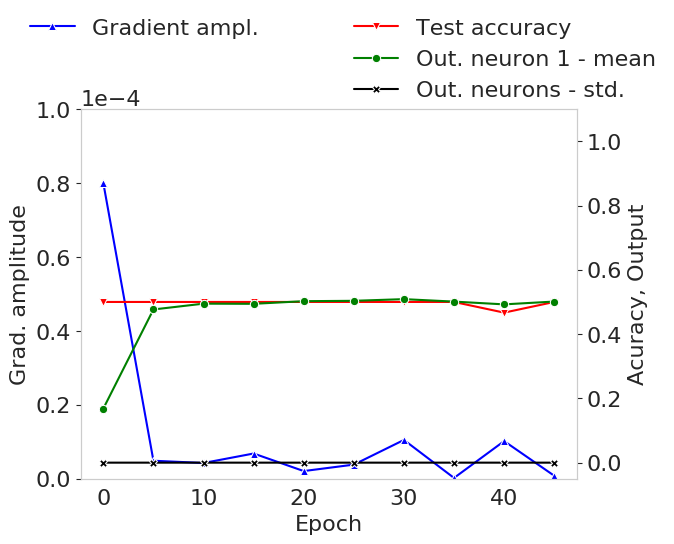}  
  \caption{Fully conflicting bundle \\ training with balanced dataset}
  \label{fig:experiment_threshold_b}
\end{subfigure}
\begin{subfigure}{.33\textwidth}
  \centering
  \captionsetup{justification=centering}
  \includegraphics[width=.99\linewidth]{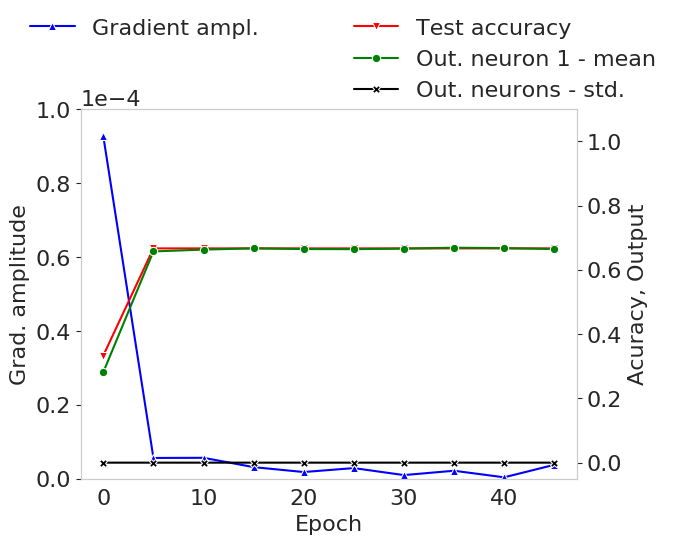}  
  \caption{Fully conflicting bundle \\ training with imbalanced dataset}
  \label{fig:experiment_threshold_c}
\end{subfigure}

\caption{Experiment with a toy dataset and manually initialized weights to produce a fully conflicting bundle for balanced and imbalanced datasets and to check if gradients resemble white noise.}
\label{fig:experiment_threshold}
\end{figure*}

\paragraph{Training.} We refer to networks that are built only with fully connected layers as fully connected networks, networks as introduced by \citet{residual_neural_networks} residual neural network (ResNet) and we call the same network without residual connections VGG net. State-of-the-art datasets \cite{imagenette, cifar10, mnist, svhn}, and random data augmentation is used to evaluate conflicting bundles under real conditions: We normalize and randomly crop, flip (except for Mnist) and adapt the brightness of images. ReLU activations are used and therefore weights are initialized with the HE initializer \cite{he_initializer} to avoid vanishing or exploding gradients. To minimize the cross-entropy loss we use the state-of-the-art optimizer Ranger (RAdam \cite{radam} + Lookahead \cite{lookahead}) with a mini-batch size of $64$, a learning rate of $0.001$, weight decay of $0.01$. ResNets and VGG nets are trained for $120$ epochs and the fully connected networks for $50$ epochs. The source uses TensorFlow Version 2.2.0 \cite{tensorflow} and is available on GitHub \footnote{\url{https://github.com/peerdavid/conflicting-bundles}}. Training is executed on a multi-GPU cluster, one GPU is used to measure conflicting bundles. All experiments are also implemented and designed to be executable on smaller systems with a single GPU.

\paragraph{Evaluation.} The test accuracy is averaged over the last $5$ epochs to exclude outliers. We estimate conflicting bundles through a random subset of the training set $\mathcal{B} \subseteq \mathcal{X}$ with $|\mathcal{B}| = 2048$ to speed up computations and we found empirically that this number of samples is large enough to represent the bundle entropy. To calculate bundles we created a vectorized function that iterates only once overall $x \in \mathcal{B}$ such that this calculation can also be done on small hardware setups with e.g. only one GPU. Therefore, the complexity to evaluate all layers is $O(|\mathcal{B}| \times L)$. It is important to mention that examples of the training data and not the test data are used to measure conflicting training bundles, as those examples are used to adjust the weights of the network.

\subsection{Training with a fully conflicting bundle} \label{sec:experimental_fully}

\begin{figure}[t]
  \centering
  \captionsetup{justification=centering}
  \includegraphics[width=.6\linewidth]{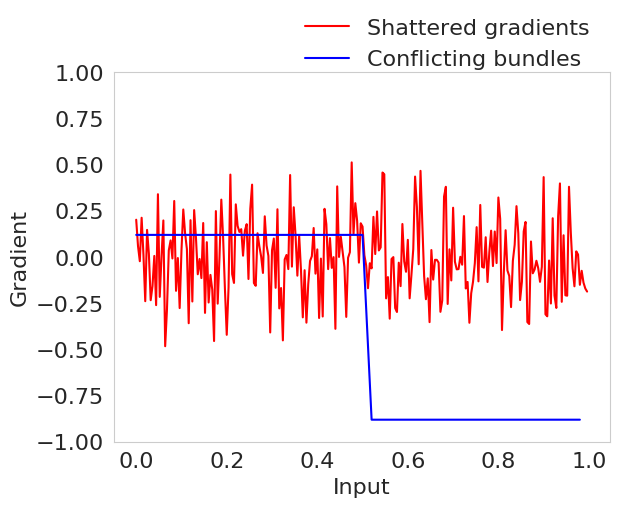}  
  \caption{White noise of shattered gradients \cite{shattered_gradients} compared with gradients of the conflicting training bundle problem.}
  \label{fig:experiment_threshold_shattered}
\end{figure}

\begin{figure*}[t]
\begin{subfigure}{.33\textwidth}
  \centering
  \captionsetup{justification=centering}
  \includegraphics[width=.85\linewidth]{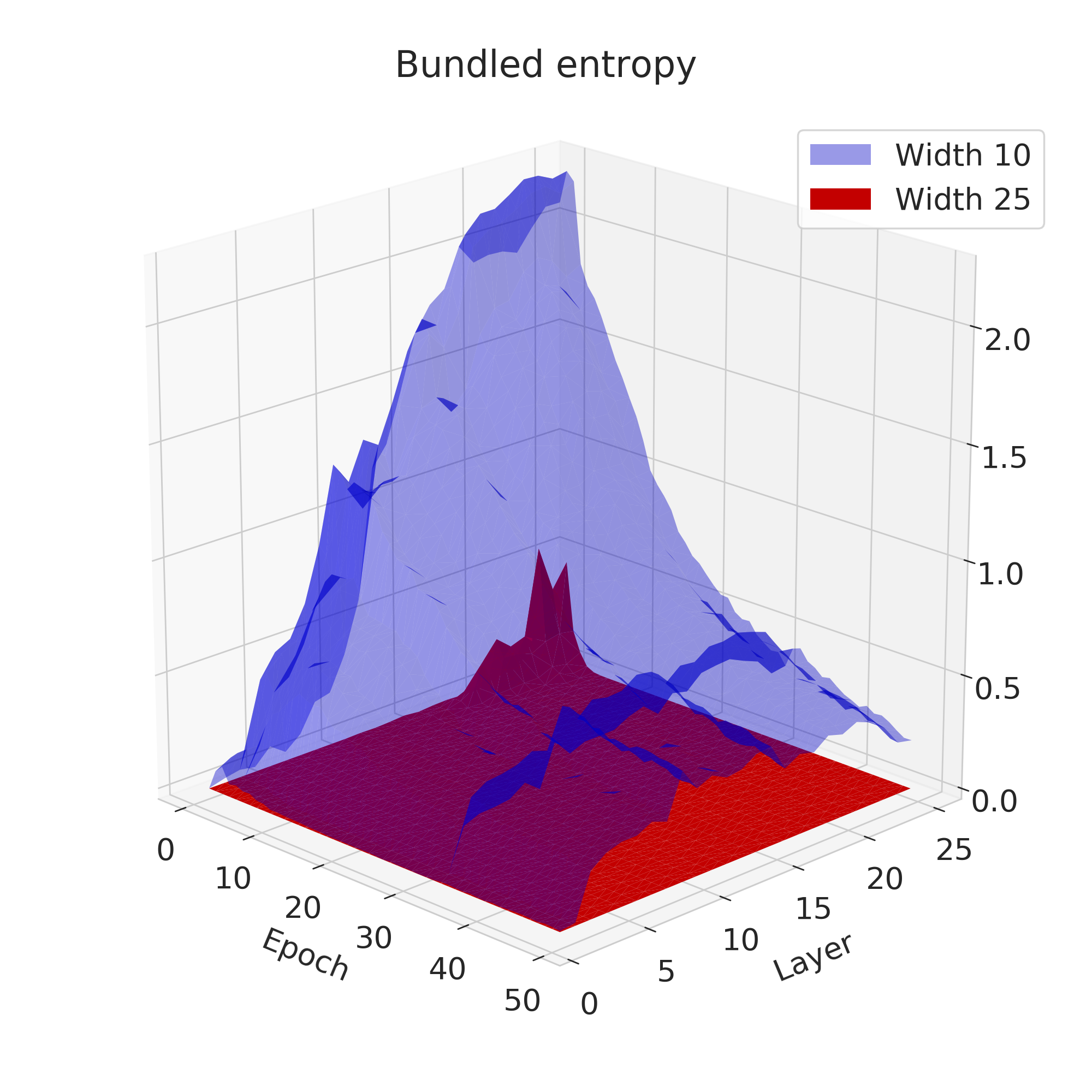}
  \caption{$H^l(t)$ evaluated for each layer $l$ and epoch $t$ for a network with $L=25$.}\label{fig:experiment_fnn_3d_a}
\end{subfigure}
\begin{subfigure}{.33\textwidth}
  \centering
  \captionsetup{justification=centering}
  \includegraphics[width=.85\linewidth]{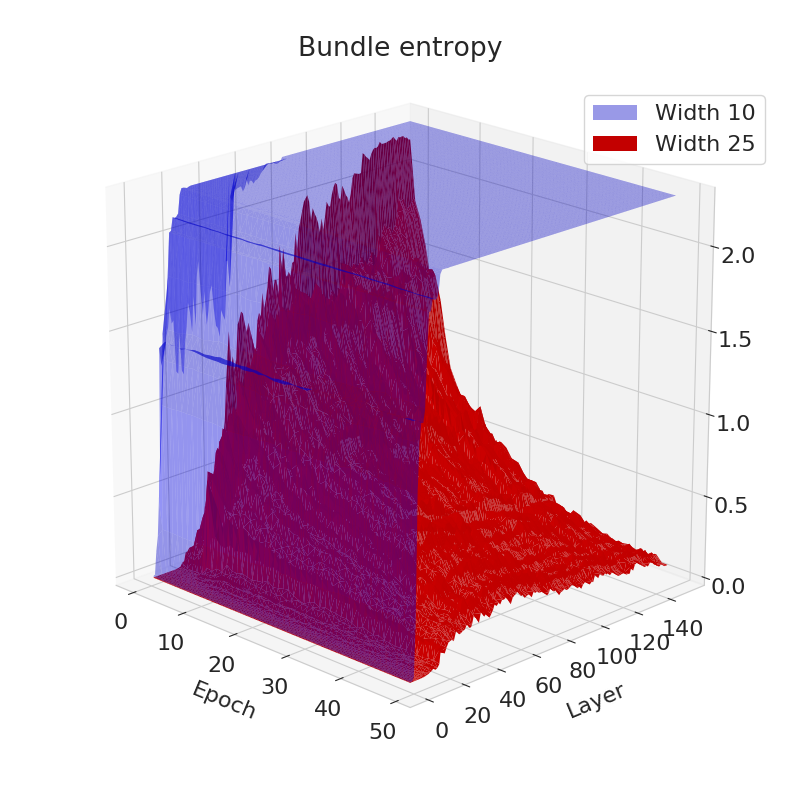}
  \caption{$H^l(t)$ evaluated for each layer $l$ and epoch $t$ for a network with $L=150$.}\label{fig:experiment_fnn_3d_b}
\end{subfigure}
\begin{subfigure}{.33\textwidth}
  \vspace{3mm}
  \centering
  \captionsetup{justification=centering}
  \includegraphics[width=.85\linewidth]{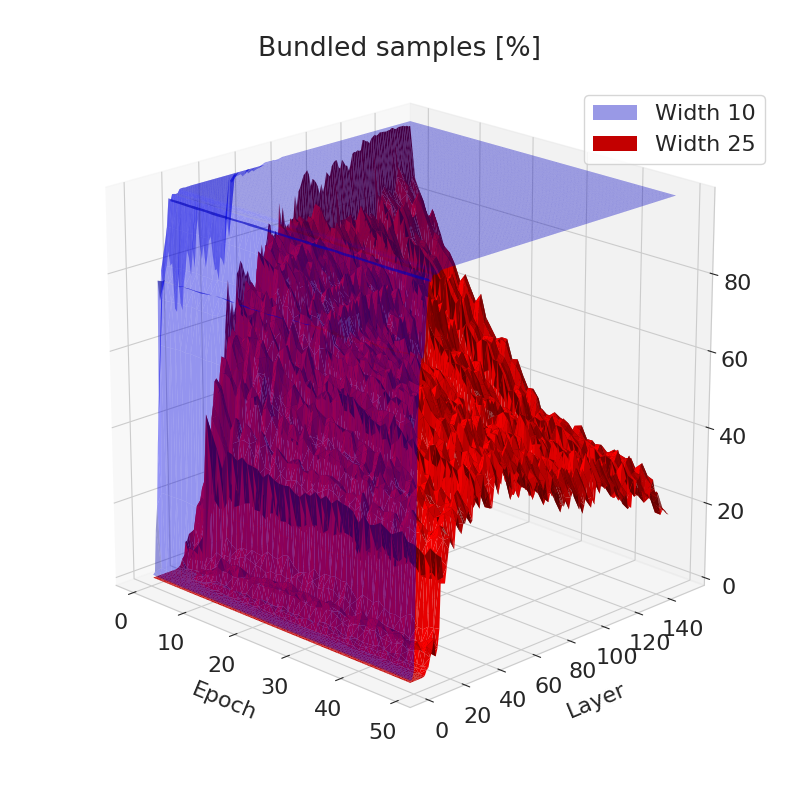}
  \caption{Number of bundles (\cref{def:bundle}) evaluated for each layer $l$ and epoch $t$ for a network with $L=150$.}\label{fig:experiment_fnn_3d_c}
\end{subfigure}
\caption{Evaluation of fully connected network w.r.t conflicting training bundles.}
\label{fig:experiment_fnn_3d}
\end{figure*}

\begin{figure}[t]
  \centering
  \captionsetup{justification=centering}
  \includegraphics[width=.70\linewidth]{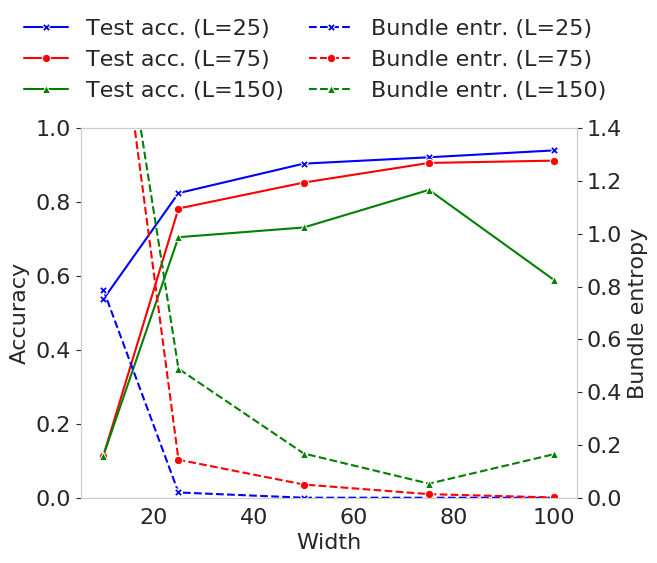}
  \caption{Test accuracy and bundle entropy $H^L$ of fully connected networks with different depth and width.}\label{fig:experiment_fnn_width}
\end{figure}


To evaluate whether training with fully conflicting bundles saddle at a region where all neurons fire with a constant value as predicted in \cref{hyp:fully}, we compare the training of a neural network with and without a fully conflicting bundle under controlled settings. The fully conflicting bundle is produced in a setup where the conditions can be tightly controlled through manual weight initialization. We use a toy dataset with two classes (class one if $x_i < 0.5$, class two otherwise) and $1$k training examples for a network with ReLU activations, two layers and two neurons per layer followed by softmax and cross-entropy loss function. The results are shown in \cref{fig:experiment_threshold}.

At first, we confirmed in \cref{fig:experiment_threshold_a} that this network can solve the problem with a training accuracy of $\approx 1.0$ if weights are initialized such that no fully conflicting bundle is produced. For the training shown in \cref{fig:experiment_threshold_b} we initialized weights such that a fully conflicting bundle is produced. It can be seen that for this weight configuration \emph{each output neuron fires with a constant value of $1 / N_c = 0.5$} after $50$ epochs as predicted by \cref{hyp:fully}. In \cref{fig:experiment_threshold_c} the network is trained under fully conflicting bundles together with an imbalanced dataset such that $66\%$ of the training examples are of class one and $33\%$ are of class two. As claimed in \cref{sec:theory_fully}, each neuron reflects the imbalance of the dataset because neuron one fires with a constant value of $0.66$. The gradient amplitude of the training without conflicting bundles (\cref{fig:experiment_threshold_a}) and with conflicting bundles (\cref{fig:experiment_threshold_b}, \cref{fig:experiment_threshold_c}) is similar at the beginning of the training process, ruling out the gradient vanishing or exploding problem. Also, the networks have only two layers such that this problem would not arise. We also compared the shattered gradient problem with the gradients of fully conflicting training bundles in \cref{fig:experiment_threshold_shattered} and can rule out that both problems are the same because \emph{gradients of conflicting bundles are highly correlated with its input}.

We executed this first experiment under controlled settings and therefore we continue with the HE initializer \cite{he_initializer}, larger fully connected networks and the Mnist dataset.

\subsection{Fully connected networks} \label{sec:experiment_fnn}

The test accuracy of more than a hundred fully connected networks together with the conflicting boundary is already shown in \cref{fig:experiment_fnn_first_conflicting_layer}. In this section, we study the bundle entropy and the number of bundles that occur during training in more detail by analyzing each training epoch and layer of different fully connected networks trained on Mnist. Results are shown in \cref{fig:experiment_fnn_3d} and \cref{fig:experiment_fnn_width}.

In \cref{fig:experiment_fnn_3d_a} we can see the training of two shallow networks with width $10$ and $25$. First of all, the number of conflicts mainly increases as we ascend the hierarchy of the network indicating that subsequent layers can not fully solve conflicts. Therefore, it can be concluded that the \emph{number of conflicting training bundles increases as the depth of the network increases}. Comparing width $10$ and width $25$ of \cref{fig:experiment_fnn_3d_a} shows that \emph{conflicts occur much earlier in the architecture if the dimensionality of hidden features is smaller}. This explains the pattern of the test-accuracy shown in \cref{fig:experiment_fnn_first_conflicting_layer}. Also, if conflicting bundles occur during training, \emph{conflicts can only slowly be resolved as the training proceeds}. This motivates why we are able to create the boundary of \cref{fig:experiment_fnn_first_conflicting_layer} already at the beginning of the training. Two deeper networks with $150$ layers are analyzed in \cref{fig:experiment_fnn_3d_b} and \cref{fig:experiment_fnn_3d_c}. For both widths conflicting bundles occur during training, but for a width of $10$ it can be seen that a fully conflicting bundle occurs because $H^l(t) \approx 2.3$. Therefore, conflicts are never resolved and we also confirmed that the accuracy is not better than chance as claimed in \cref{hyp:fully}. Interestingly, if we compare \cref{fig:experiment_fnn_3d_a} and \cref{fig:experiment_fnn_3d_b} it can be concluded that \emph{for a fixed width, the position where conflicting layers occur is similar between different depths}. For example the first conflicting layer for width $25$ is layer $19$ for both, $L=25$ and $L=150$. Therefore, it is sufficient to only evaluate the deepest network to find the conflicting boundary in \cref{fig:experiment_fnn_first_conflicting_layer}. The correlation of the bundle entropy and test accuracy is evaluated in \cref{fig:experiment_fnn_width}: It can be seen that \emph{the bundle entropy $H^L$ is negatively correlated with the test accuracy}. This further supports our \cref{hyp:partially} that conflicting bundles worsen the test accuracy of neural networks.

One natural question that arises from this section is, whether conflicts also occur for computer vision tasks and convolutional layers because hidden features are high dimensional.

\subsection{VGG nets}\label{sec:experiment_vgg}

\begin{figure*}[t]
\begin{subfigure}{.33\textwidth}
  \vspace{1mm}
  \centering
  \captionsetup{justification=centering}
  \includegraphics[width=.97\linewidth]{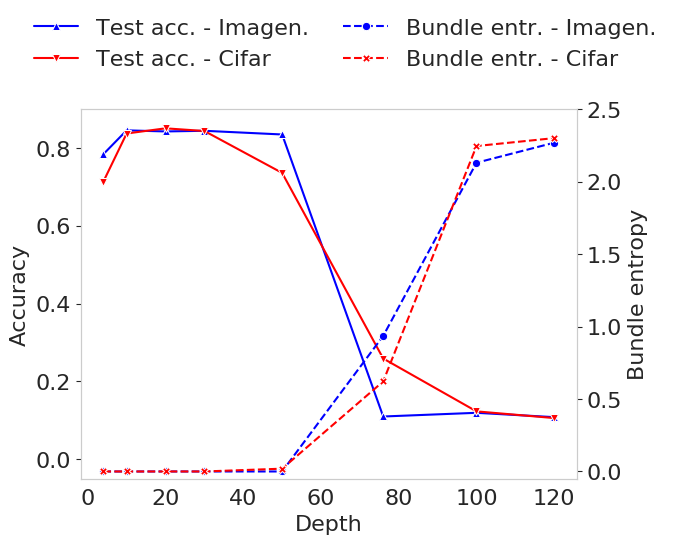}
  \caption{Test accuracy and bundle entropy $H^L$ for different num. of layers without residual connections}\label{fig:experiment_vgg_a}
\end{subfigure}
\begin{subfigure}{.30\textwidth}
  \centering
  \captionsetup{justification=centering}
  \includegraphics[width=.93\linewidth]{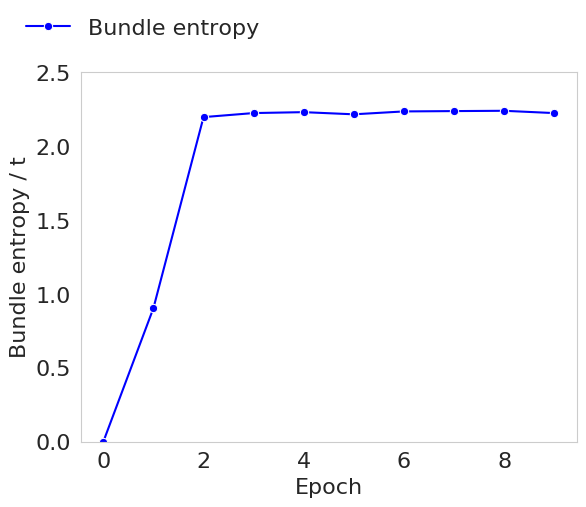}
  \caption{$H^L(t)$ for a network without \\ residuals and $100$ layers}\label{fig:experiment_vgg_b}
\end{subfigure}
\begin{subfigure}{.33\textwidth}
  \centering
  \captionsetup{justification=centering}
  \includegraphics[width=.99\linewidth]{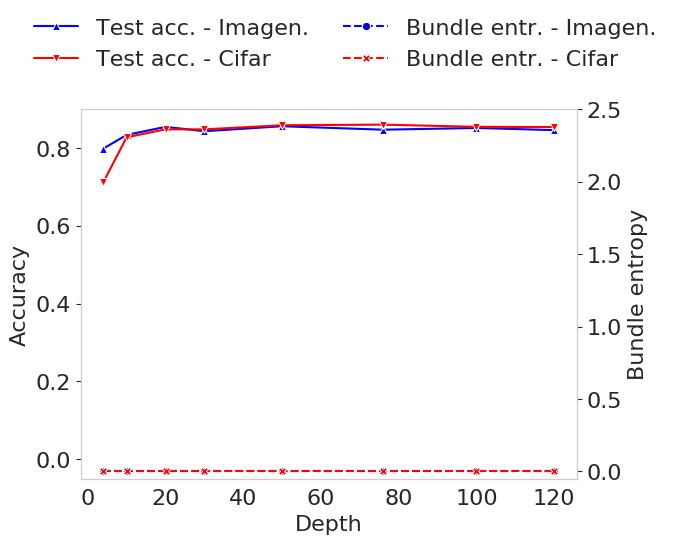}
  \caption{Test accuracy and bundle entropy $H^L$ for different num. of layers with residual connections}
  \label{fig:experiment_vgg_c}
\end{subfigure}
\caption{Performance and bundle entropy of different VGG nets and residual neural networks.}
\label{fig:experiment_vgg}
\end{figure*}
In this section, conflicting bundles are evaluated for different VGG nets and different datasets to further support \cref{hyp:fully} and \cref{hyp:partially} experimentally. The network is trained without residual connections and the results are shown in \cref{fig:experiment_vgg_a}. The bundle entropy of each time step $H^L(t)$ for a network with $100$ layers trained on Imagenette is shown in \cref{fig:experiment_vgg_b}.

For small networks with only four layers, the model suffers from underfitting and therefore the test accuracy is lower. From $10$ to $40$ layers, the test accuracy is largest and the bundle entropy is also zero indicating no conflicting bundles during training. After $40$ layers for Cifar (and $60$ layers for Imagenette), the bundle entropy $H^L$ increases as the depth increases. The test accuracy decreases as the bundle entropy increases which supports \cref{hyp:partially}. For $120$ layers the entropy is $\approx 2.3$ (max. possible entropy for $10$ classes is $2.3$) indicating a fully conflicting bundle and therefore the accuracy is also not better than chance as hypothesized in \cref{hyp:fully}. Previous work already reported that the test accuracy decreases when training very deep convolutional networks \cite{shattered_gradients, residual_neural_networks, highway_networks}. We are now able to relate this behavior additionally to conflicting bundles. Another interesting observation is that weights are initialized correctly because at the beginning of the training the bundle entropy is zero as shown in \cref{fig:experiment_vgg_b} and conflicting bundles occur after the first epochs of training. We conclude that weight initialization is very important, but \emph{conflicts can also occur even when weights are initialized correctly and the probability for conflicting layers increases with each layer that is added to the architecture also for very high dimensional features such as images}. 

It is well known that residual connections reduce the shattered gradient problem and we have seen in \cref{sec:experimental_fully} and  \cref{sec:experiment_fnn} that shattered gradients are different than conflicting bundles. Therefore the question of whether residual connections also solve the conflicting bundle problem is still an open one and analyzed next.

\subsection{Residual neural networks}\label{sec:experiment_residual}
\begin{table*}[t]
  \caption{Comparison of test accuracy, memory consumption (based on checkpoint size) and inference time.}
  \label{tbl:auto_tune}
  \centering
  \begin{tabular}{llc|ccc}
    \hline
    Dataset & Name     & Layers     & Accuracy [\%] & Mem. [MB] & Time / Step [ms] \\
    \hline
    Imagenette & ResNet & $50$ & $85.6 \pm 0.8$ & $116$ & $310$    \\
               & \textbf{Auto-tune} & $\pmb{24 \pm 1.6}$ & $\pmb{84.7 \pm 0.1}$ & $\pmb{63 \pm 6.0}$ & $\pmb{200 \pm 0.0}$ \\
    \hline
    Cifar & ResNet & $76$ & $85.7 \pm 0.4$ & $178$ & $130$   \\
          & \textbf{Auto-tune} & $\pmb{19 \pm 0.9}$ & $\pmb{85.3 \pm 0.4}$ & $\pmb{53 \pm 1.0}$ & $\pmb{48 \pm 2.5}$ \\
    \hline
    Svhn & ResNet & $50$ & $95.4 \pm 0.1$ & $116$ & $90$    \\
               & \textbf{Auto-tune} & $\pmb{19 \pm 0.9}$ & $\pmb{95.1 \pm 0.1}$ & $\pmb{54 \pm 0.5}$ & $\pmb{46 \pm 1.8}$    \\
    \hline
    Mnist & ResNet & $50$ & $ 99.4 \pm 0.0$ & $116$ & $88$ \\
               & \textbf{Auto-tune} & $\pmb{ 17 \pm 1 }$ & $\pmb{ 99.3 \pm 0.5 }$  & $\pmb{54 \pm 2.1}$ & $\pmb{ 40 \pm 0.0}$    \\
    \hline
  \end{tabular}
\end{table*}
\david{[Done] The proof is confusing for reviewers.}
To evaluate whether residual connections solve conflicts, we trained residual networks with different depths on different datasets. It can be seen in \cref{fig:experiment_vgg_c} that the bundle entropy is zero for all network depths and datasets, thus it can be concluded that \emph{conflicts are solved by residual connections and therefore, the training of very deep convolutional networks is possible}. Also, the accuracy is high for all different residual networks which further supports \cref{hyp:partially}. \citet{residual_neural_networks} already showed that residual networks are easier to optimize and our conflicting bundle theory explains this fact from a new perspective.

We further evaluate \emph{how} residual connections bypass conflicting layers from a theoretical viewpoint: Let's consider layer $l$ producing conflicts for $x_i$ and $x_j$ without any residual connection. We call this the intermediate conflicting output $d^{(l)}$ (which is the same for $x_i$ and $x_j$ as it is assumed to be conflicting). The output for inputs $x_i$ and $x_j$ of the layer which adds a residual connection $r^{(l)}(x)$ is $a^{(l+1)}(x) = r^{(l)}(x) + d^{(l)}$. The residual $r^{(l)}(x)$ is the identity mapping \cite{residual_neural_networks} and therefore it can be shown that the function $a^{(l+1)}(x)$ is bijective for inputs $x_i$ and $x_j$. \Cref{def:bundling} is violated because $a^{(l+1)}_i \neq a^{(l+1)}_j$ and the conflict is resolved. This analysis shows how residual connections solve conflicts. But, it is also important to mention that this analysis not proved the absence of conflicts in general. For example, if the intermediate layer is exactly the negative identity function, the output $a^{(l+1)}(x)$ is constantly zero and therefore, conflicting. On the other hand, \citet{residual_neural_networks} already showed that it is hard to learn the identity function and therefore we assume that the aforementioned special case never happens in practise. This is backed by the results shown in \cref{fig:experiment_vgg_c} because the bundle entropy is always zero.

Another important insight of this theoretical analysis is that \emph{conflicting layers that are bypassed with residual connections represent only a linear mapping, because $a^{(l+1)}(x) = r^{(l)}(x) + d^{(l)}$}. Therefore, we assume that the test accuracy of correctly pruned networks (w.r.t conflicting layers) is not significantly worse than its residual counterpart which is evaluated next.

\subsection{Auto-tuning the network depth} \label{sec:experiment_auto_tune}
The pruning algorithm that we propose searches for the conflicting boundary as shown in \cref{fig:experiment_fnn_first_conflicting_layer} for a given input dimensionality and prunes afterwards the network to ensure that no conflicting bundles occur during training. This is implemented as follows: First, the largest network from \cref{sec:experiment_vgg} with $120$ layers is trained for at least one epoch (see \cref{fig:experiment_vgg_b}). Then the first layer $l$ with  $H^l(t) > 0$ and all subsequent layers of the same type (specified in \cite{residual_neural_networks}) are removed from the architecture following the findings of \cref{sec:experiment_fnn}. Due to this pruning, also the dimensionality between two layers changes and therefore we restart the training with the new pruned architecture rather than continuing the training to avoid dimensionality problems. A restart additionally helps in terms of accuracy, because the network is trained for one epoch with conflicting bundles such that weights of the network are adjusted into wrong directions. This process is repeated until no conflicting layer can be found and the network is successfully trained.

To provide a fair comparison, we compare the pruned network found by the auto-tune algorithm against the ResNet which produced the highest accuracy (\cref{sec:experiment_residual}). To evaluate whether a similar accuracy can be achieved for other datasets, the Svhn \cite{svhn} and Mnist \cite{mnist} dataset are included and trained on ResNet-$50$ as this depth produced good results in \cref{sec:experiment_residual}. Note that each experiment is executed three times and the mean and standard deviation are reported in \cref{tbl:auto_tune}. From our experimental evaluation, we observed that the auto-tune algorithm changed the architecture at most three times before the final pruned network was found and the architecture was changed after the first and before the second epoch. Therefore, this automatic depth selection process is computationally very efficient as it took only three epochs of training to select the architecture. The number of layers found by the auto-tune algorithm is within the optimal region w.r.t. test accuracy as shown in \cref{fig:experiment_vgg_a} and the \emph{depth increases as its input dimensionality increases} similar to the findings from \cref{sec:experiment_fnn}. Interestingly, the number of layers also corresponds with the studies of \citet{residual_behave_shallow} which shows that paths in ResNets are only 10-34 layers deep. From \cref{tbl:auto_tune} it can also be concluded that \emph{the test accuracy of networks without conflicting layers are competitive when compared to its residual counterpart, but the inference time and the memory consumption is drastically reduced because fewer layers are used.} 


\section{Discussion and future work}\label{sec:discussion}

\david{Discuss why conflicting bundles occur.}
In this paper, we have shown that conflicting training bundles decrease the test accuracy of trained models or lead to networks that can not be trained at all. We have shown that the number of conflicting training bundles increases, as the dimensionality of hidden features decreases or the depth of the network increases. We demonstrated that this also happens for computer vision tasks where hidden features are very high dimensional and we proved that residual connections solve conflicts. The mapping from an input to its output of a conflicting layer bypassed with a residual connection is only linear and therefore, the pruning of those layers should lead to competitive test accuracy when compared to its residual counterpart. We introduced a novel NAS algorithm to remove conflicting layers and confirmed that the accuracy is maintained while the computational power and memory consumption is drastically reduced.

\david{Add floating point idea, data augmentation and sampling idea}
For future work, it would be interesting to study how conflicting layers can be directly fixed rather than removed from the architecture. The analysis of \cref{sec:experiment_residual} suggests to use a different approach than residual connections, because they force a linear mapping from inputs to outputs. Another promising direction of research would be to study if novel data-augmentation strategies or intelligent sampling of mini-batches could help to avoid conflicting bundles during training. Another interesting topic to study would be to evaluate whether a coarser resolution of floating point representations bundles samples with a higher probability. To correctly evaluate this hypothesis, it must be considered that the resolution $\gamma$ is already included in the metric that we propose because this biases the metric into the hypothesized direction.



\section*{Acknowledgments}
\david{Is this ok, what do you think?}
We acknowledge all members of the IIS research group, the European Union’s Horizon 2020 program for the grant agreement no. 731761 (IMAGINE) and DeepOpinion for the opportunity to continue with this research in the future.

{\small
\bibliographystyle{plainnat}
\bibliography{egbib}
}

\end{document}